\documentclass[runningheads]{llncs}
\usepackage[title]{appendix}
\usepackage{graphicx}
\usepackage{comment}
\usepackage{amsmath,amssymb} 
\usepackage{dsfont}
\usepackage{color}
\usepackage{booktabs}
\usepackage{colortbl}
\usepackage[table]{xcolor}
\usepackage{makecell}
\usepackage{siunitx,booktabs,etoolbox}
\usepackage[capitalise]{cleveref}
\usepackage[misc]{ifsym}

\usepackage{amsmath,amssymb} 
\usepackage{dsfont}
\usepackage{color}
\usepackage{booktabs}
\usepackage{colortbl}

\usepackage{makecell}
\usepackage{xspace}
\usepackage{scalerel}
\usepackage{amsmath}
\usepackage{algorithm,algpseudocode}

\algnewcommand{\algorithmicforeach}{\textbf{for each}}
\algdef{SE}[FOR]{ForEach}{EndForEach}[1]
  {\algorithmicforeach\ #1\ \algorithmicdo}
  {\algorithmicend\ \algorithmicforeach}

\usepackage{xspace}
\usepackage{scalerel}
\usepackage{amsmath}

\makeatletter
\DeclareRobustCommand\onedot{\futurelet\@let@token\@onedot}
\def\@onedot{\ifx\@let@token.\else.\null\fi\xspace}

\def\eg{\emph{e.g}\onedot} 
\def\ie{\emph{i.e}\onedot} 
\def\cf{\emph{c.f}\onedot} 
 \def\vs{\emph{vs}\onedot}
 
\def\etal{\emph{et al}\onedot}
\makeatother

\begin{document}
\pagestyle{headings}
\mainmatter
\def\ECCVSubNumber{438}  


\title{Many-shot from Low-shot: Learning to Annotate using Mixed Supervision for Object Detection} 

\titlerunning{Learning to Annotate Mixed Supervision for Object Detection}
%
\author{Carlo Biffi\inst{1} \and
Steven McDonagh\inst{1} \and
Philip Torr\inst{3} \and 
Ale\v{s} Leonardis\inst{1} \and \\
Sarah Parisot \inst{1,2}}
\authorrunning{C. Biffi et al.}
%
\institute{Huawei Noah's Ark Lab \and
Mila Montr\'eal \and  
University of Oxford
}
\maketitle


\begin{abstract}
Object detection has witnessed significant progress by relying on large, manually annotated datasets. Annotating such datasets is highly time consuming and expensive, which motivates the development of weakly supervised and few-shot object detection methods. However, these methods largely underperform with respect to their strongly supervised counterpart, as weak training signals \emph{often} result in partial or oversized detections. Towards solving this problem we introduce, for the first time, an online annotation module (OAM) that learns to generate a many-shot set of \emph{reliable} annotations from a larger volume of weakly labelled images. Our OAM can be jointly trained with any fully supervised two-stage object detection method, providing additional training annotations on the fly. This results in a fully end-to-end strategy that only requires a low-shot set of fully annotated images. The integration of the OAM with Fast(er) R-CNN improves their performance by $17\%$ mAP, $9\%$ AP50 on PASCAL VOC 2007 and MS-COCO benchmarks, and significantly outperforms competing methods using mixed supervision.

\end{abstract}

\section{Introduction}









Object detection is an essential building block of many computer vision systems \cite{zhao2019object}. State-of-the-art (SOTA) methods mainly rely on large scale datasets with manually annotated bounding boxes to train fully supervised CNN-based models~\cite{girshick2015fast,ren2015faster,redmon2016you,lin2017focal,cai2019cascade}. However, the prohibitive cost and time requirements associated with data annotation reduce the applicability of SOTA detection models in real life scenarios. This has motivated research on object detection strategies with reduced data annotation requirements. Amongst the most popular low data regimes, we distinguish Weakly Supervised Object Detection (WSOD), which aims to train object detectors using only image-level annotations~\cite{bilen2016weakly,tang2018pcl,wan2019c,arun2019,zeng2019wsod2}, and Few-Shot or Low-Shot Object Detection (FSOD/LSOD), training supervised models with only a handful of training examples on all (LSOD) or only a subset of novel test classes (FSOD)~\cite{kang2019few,yan2019meta,dong2018few}. FSOD and in particular WSOD have been the focus of a large body of work with innovative strategies obtaining promising performance. Nonetheless, these models typically fall far short of their strongly supervised counterparts. Numerical performance gaps are attributed to the low quality of bounding-box annotations produced, \eg by WSOD methods, that often manifest as partial or oversized boxes. Such results are not reliable enough for use in real-world scenarios and can be observed to cause deterioration of detection performance when used in fully supervised models training. This can be attributed to weak training signals requiring very large and curated datasets (WSOD) or very representative and carefully selected annotated examples (FSOD). 



\begin{figure}[t]
\centering
\includegraphics[width=0.95\linewidth]{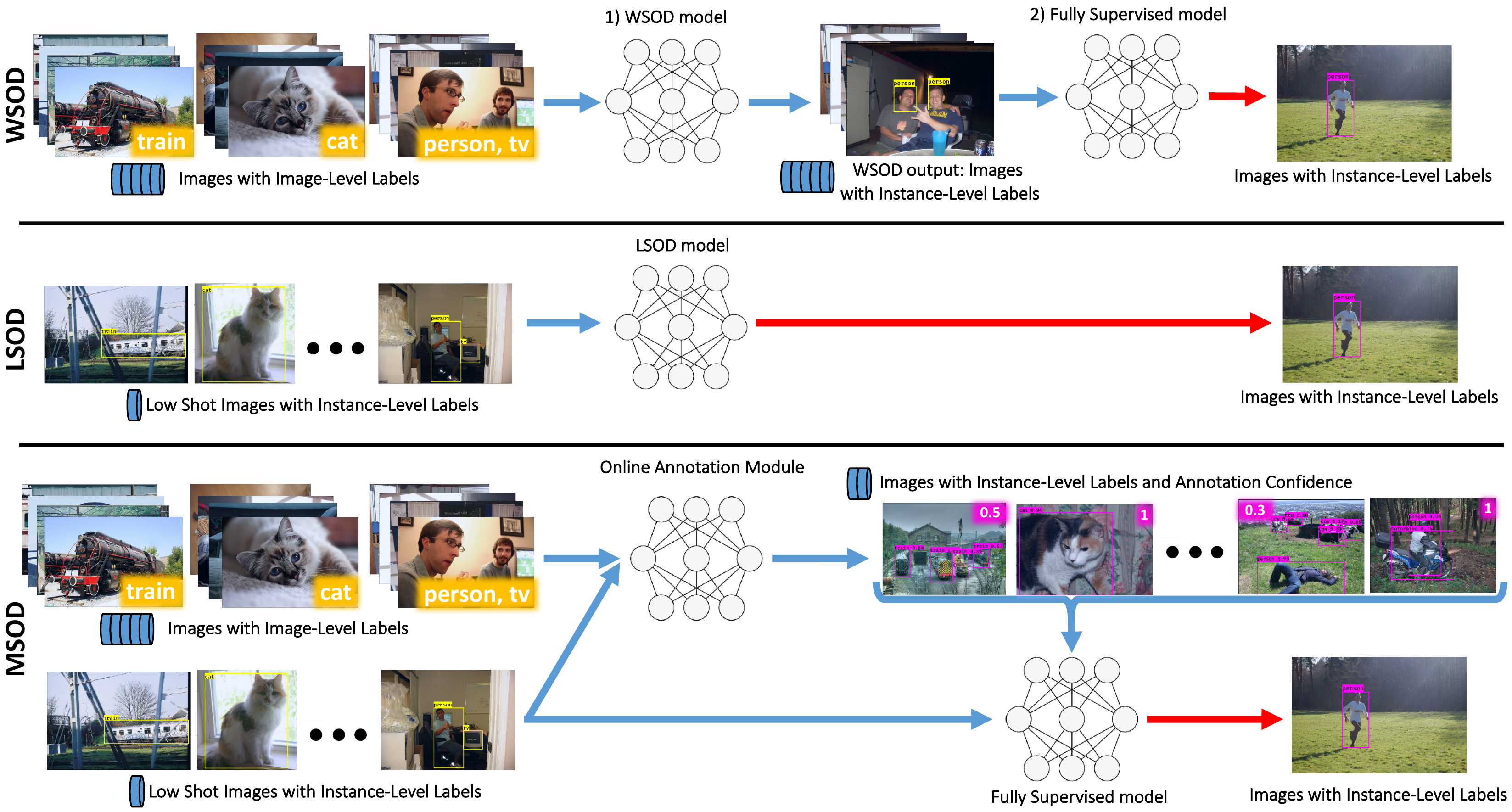}
\caption{Weak and low data detection strategies and our proposed mixed supervision-based setting. First row: Weakly Supervised Object Detection (WSOD) models learn to annotate images with image-level annotations which are then used to train fully supervised models. WSOD annotations are often partial or oversized, resulting in poor detector performance. Second row: 
Few-shot or low-shot object detection (LSOD) trains models on only a handful of training examples. Research mainly focuses on situations where only a specific subset of novel classes have limited training data.
Bottom row: Mixed Supervision for Object Detection (MSOD) combines a low shot set of images containing object annotations with a large volume of images comprising only image-level annotations. We train an online annotation module to generate a many-shots set which, at the same time, is used to train a fully supervised model.}
\label{fig:low_data_strats}
\end{figure}

To address the aforementioned challenges, we focus on a recent training paradigm relying on Mixed Supervision for  Object Detection  (MSOD)~\cite{pan2019low,fang2020low}. The distinction between this protocol and the previously introduced weak and low data settings is illustrated in Fig.~\ref{fig:low_data_strats}. The objective of MSOD is to exploit and combine the complementary advantages provided by 
WSOD and LSOD; weak (image-level) supervision affords the construction of large databases with minimal effort, while low-shot supervision provides information rich, fully annotated ground truth examples. The MSOD paradigm has, only very recently, been initially investigated in two related works. Fang \etal~\cite{fang2020low} propose a cascaded architecture yielding performance competitive with fully supervised counterparts yet using a significant fraction of the full training data to achieve comparable performance. Pan \etal~\cite{pan2019low} use low-shot examples to refine bounding box annotations obtained from a pre-trained WSOD model~\cite{tang2017multiple}, resulting in a method intrinsically linked to the performance and drawbacks of WSOD techniques. 
In this work, we approach the MSOD scenario from a different angle. Due to the sparsity of rich training information provided, we expect a MSOD model to output annotations of variable quality, especially for images containing crowded scenes or objects with appearance substantially dissimilar to the training data. In contrast to existing MSOD models we introduce an Online Annotation Module (OAM), trained with mixed supervision, that can be used in conjunction with any two-stage fully supervised object detection method to improve its performance (\eg Fast(er) R-CNN family~\cite{girshick2015fast,ren2015faster}). Our OAM generates, on the fly, additional reliable automated annotations obtained from a larger set of weakly annotated images (containing only image-level class labels). Furthermore, we exploit prediction stability to reason about annotation reliability resulting in associated confidence scores. Generated annotations are used to train, concurrently to the OAM, a fully supervised detector that shares the same encoding features. This produces an intrinsic training curriculum for the standard detector model; only simple images, labelled with high confidence will be presented to the model at the outset. Compared to previous MSOD work, our OAM strategy provides increased robustness against mislabeled crowded and ambiguous training images as only confident MSOD annotations are exploited for fully supervised training. Furthermore, our joint MSOD and fully supervised training provides intrinsic regularisation for both tasks, allowing the learning of higher quality and more discriminative feature extractors. 

Experiments show that our strategy allows effective training of standard detection algorithms with only minimal annotation requirements and significantly outperforms WSOD and competitive MSOD approaches on PASCAL VOC 2007 and MS-COCO benchmarks. Additionally, we report competitive performance in comparison to fully supervised alternatives, illustrating the ability of our OAM to annotate a many-shot set of (weakly labelled) images 
that can be leveraged to improve the fully supervised model performance.

In summary, we propose a new direction using Mixed Supervision for Object Detection (MSOD). Our main contributions are the following: 

\begin{itemize}
    
    \item We introduce a novel Online Annotation Module (OAM), trained using mixed supervision. This module allows expansion of the low-shot training set of fully annotated images by generating \emph{reliable} annotations from a larger volume of weakly labelled images.

    \item Training our OAM concurrently with any two-stage object detection model introduces a strategy for object detection performance improvement due to the generated annotation. We report on the benefits of intrinsic regularisation afforded to both tasks when common encoding features are shared.
    
    
    
    
    \item The integration of the OAM with Fast(er) R-CNN improves their performance by $17\%$ mAP, $9\%$ AP50 on PASCAL VOC 2007 and MS-COCO benchmarks, and significantly outperforms MSOD approaches.

    
    
    
    
    
    
    
    
\end{itemize}


\section{Related work}

\noindent\textbf{Weakly Supervised Object Detection.}
A large body of recent work, considering WSOD, couples CNN feature extractors with Multiple Instance Learning (MIL) frameworks, thus casting weakly supervised object detection as a multi-label classification problem. Each image is typically represented as a bag of pre-computed proposals (\eg Selective Search~\cite{uijlings2013selective}, Edge Boxes~\cite{zitnick2014edge}, etc.) and the objective is to identify proposals that are most relevant for bag classification ~\cite{bilen2016weakly,tang2018pcl,wan2019c,zeng2019wsod2}. Being framed as a classification task, MIL WSOD models typically focus on proposals that comprise of either the most discriminative object parts or image regions that define the presence of an object category. They therefore struggle to detect full object extent (\eg human faces in contrast to an entire human body) or group multiple object instances of the same object within a single bounding box~\cite{zeng2019wsod2,pan2019low}. In order to address this issue, recent work has focused on bounding-box refinement strategies using cascaded refinements of MIL classifications~\cite{tang2017multiple,tang2018pcl}, using saliency maps~\cite{wei2018ts2c,zeng2019wsod2}, adopting continuation strategies~\cite{wan2019c,wan2019Pami} and modelling uncertainty~\cite{arun2019}. However, the ill-posed nature of the WSOD problem and insufficient statistics provided by the PASCAL VOC dataset (on which these approaches are usually evaluated) has lead to the development of ad-hoc training strategies and parameter sensitive methods to cope with the weak training signal, which substantially reduce generalisability across datasets. 
In this paper, we argue that including a handful of labelled samples 
yields accuracy and stability model improvements at only minimal annotation cost.
Usually, all the images annotated by MIL WSOD methods are used, in a second step, to train fully supervised models~\cite{tang2018pcl,wan2019c,zeng2019wsod2}. Further previous work has also focused on alternating between the pseudo-labelling of images and, in conjunction, training a fully supervised model~\cite{jie2017deep,dong2018few}. In this work, we generate bounding box annotations on the fly from mixed supervision and we concurrently train a fully supervised detector \emph{only} on the images annotated with high confidence.

\noindent\textbf{Few-Shot and Mixed Supervision Object Detection.}
Few-Shot Object Detection (FSOD) considers a fully supervised training set, and aims to achieve strong performance on a set of novel classes comprising of only $K$ annotated training images per class. To date only a handful of works have focused on FSOD~\cite{kang2019few,yan2019meta,karlinsky2019repmet,chen2018lstd}. Such approaches typically adapt few-shot classification techniques to the object detection setting, exploring metric learning~\cite{karlinsky2019repmet} or meta-learning~\cite{yan2019meta} strategies.
Mixed Supervision for Object Detection (MSOD) enhances a WSOD training set containing only image-level labels with a small set of fully annotated (strong) images (\eg $K$ images per class, analogous to an FSOD scenario) and aims to achieve strong performance on \emph{all} the training classes. Pan \etal recently propose BCNet~\cite{pan2019low}, which learns to refine the output of a pre-trained WSOD model using a small set of strong images. The definition of small set explored in their work ranges from $10$ shots to $20\%$ of the entire dataset (${\sim}1000$ images in PASCAL VOC 2007 training set). This approach provides a strong performance increase with respect to WSOD methods, however remains highly dependent on the original WSOD model detections as input. If detections are originally missed by the pre-trained model, the approach cannot recover. Moreover, BCNet requires the training of two independent models which makes the adaption of WSOD parameters, \ie training for new datasets, challenging. In this work, we instead propose a one-stage approach relying on an adaptive pool of annotations, updated dynamically as training progresses. EHSOD~\cite{fang2020low} and BAOD~\cite{pardo2019baod} focus on larger data regimes (\eg $10\%$ to $90\%$) and aim to reduce the data required to reach fully supervised performance using a cascaded MIL model and a student-teacher setup trained on weak and strong annotations, respectively. In contrast to all outlined methods, we propose instead to learn, and annotate on the fly, only a subset of weak images that can be labelled with high confidence. These additional samples are then used together with strong images 
to train an object detector and thus improve performance.

\section{Method}
\begin{figure*}[t]
\centering
\includegraphics[width=0.95\linewidth]{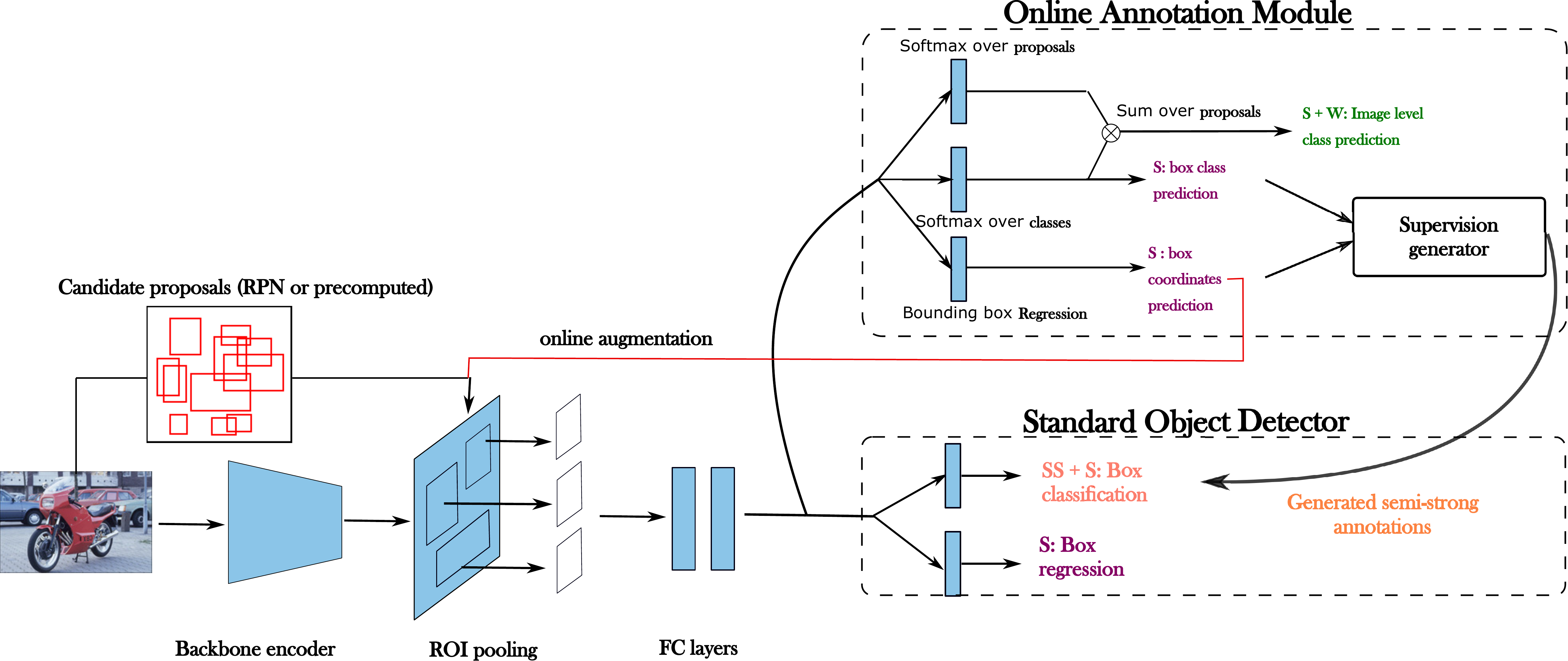}
\caption{Architecture of the proposed approach. Our model comprises two branches with a shared encoder. 
Our Online Annotation Module (OAM) is trained on weakly ($\mathcal{W}$) and strongly ($\mathcal{S}$) annotated images to generate, on the fly, confident annotations for $\mathcal{W}$ images which are added to a pool of semi-strong ($\mathcal{SS}$) images. The model's second branch uses $\mathcal{SS}$ and $\mathcal{S}$ images to train a standard fully supervised detection model.}
\label{fig:architecture_overview}
\end{figure*}

Let $\mathcal{I}$ be a set of training images annotated with image-level supervision. Under our mixed supervision paradigm, a subset of these images, $\mathcal{S} \subset \mathcal{I}$ with $|\mathcal{S}| \ll |\mathcal{I}|$, is further annotated with bounding box annotations. We refer to the images contained in $\mathcal{S}$ as \emph{strong} training images, while the images in  $\mathcal{W} = \mathcal{I} \setminus \mathcal{S}$, that have only image-level annotations, are referred to as \emph{weak} training images. An overview of our proposed method is reported in Fig.~\ref{fig:architecture_overview}. Our model comprises two branches with shared encoder backbone, and employs an ROI pooling layer to compute a fixed-length feature representation for each image bounding box proposal. The first branch of our model employs both weak and strong training images to learn an Online Annotation Module (OAM) for weak training images. The OAM generates bounding box annotations, with associated confidence scores, on the fly, for every weak training image. Annotated weak images are added to a third set of images, $\mathcal P \subset \mathcal W$, if they have been annotated with high confidence, and can be subsequently removed if their 
annotation confidence drops during training. 
Images contained in $\mathcal P$ are referred to as \emph{semi-strong} training images 
throughout the paper. The second branch of our model is designed as a standard fully supervised component and trained, 
in parallel, in an end-to-end manner using strong and semi-strong images. At testing, only the fully supervised model is used for object detection.

Given an input image, we first compute a set of $B$ candidate proposals $\{ b_r \}_{r=1}^{B}$, using either an unsupervised method (\eg Selective Search~\cite{uijlings2013selective} or Edge Boxes~\cite{zitnick2014edge}) or a Region Proposal Network (RPN)~\cite{ren2015faster}, and their associated feature vectors $\{  \boldsymbol{\xi_{r}} \}_{r=1}^{B}$. 
These feature vectors $\{ \boldsymbol{\xi_{r}} \}_{r=1}^{B}$ are obtained using a standard CNN backbone and ROI Pooling layer and provide a common input to both of our model branches: the OAM 
and the fully supervised branch.



\subsection{Online Annotation Module}
Our OAM is designed to jointly exploit weak and strong supervision in an efficient manner. It comprises three main components: 1) a joint detection module exploiting weak and strong labels in a single, common architecture to predict bounding boxes and their classes, 2) an online bounding box augmentation step that generates refined bounding box proposals, 3) a supervision generator, identifying confident annotations to be used as supervision. We next describe all three components in detail.



\noindent\textbf{Joint detection module.}
\label{sec3:lowshot_guided_mil}
Similarly to the strategy proposed in \cite{fang2020low}, we combine a multiple instance learning (MIL) type image-level classification task with a fully supervised joint classification and regression task. Our joint detection module hence comprises  three parallel, fully connected layers focusing on three different subtasks: proposal scoring, classification and regression (Fig.~\ref{fig:architecture_overview}, online annotation module block). Proposal scoring $\boldsymbol{\gamma_{\scaleto{C}{3pt}}}(c,l)$ and classification $\boldsymbol{\gamma_{\scaleto{R}{3pt}}}(c,l) \in \mathds{R}^{C \times B}$ are obtained by applying the softmax function to the output of their layers along both dimensions, independently, (classes for $\boldsymbol{\gamma_{\scaleto{C}{3pt}}}$, proposals for $\boldsymbol{\gamma_{\scaleto{R}{3pt}}}$). After this operation, $\boldsymbol{\gamma_{\scaleto{C}{3pt}}}(c,l)$ 
represents the probability that the $l$-th proposal belongs to class $c$, while $\boldsymbol{\gamma_{\scaleto{R}{3pt}}}(c,l)$ represents the 
proportional contribution that proposal $l$ provides to the image being classified as class $c$. Following \cite{bilen2016weakly}, these layers are trained by exploiting the image-level supervision of both strong and weak images. In particular, a proposal score $\boldsymbol{\phi_{\scaleto{P}{3pt}}} = \boldsymbol{\gamma_{\scaleto{C}{3pt}}} \odot  \boldsymbol{\gamma_{\scaleto{R}{3pt}}}$, per class, is obtained by combining them where $\odot$ is a Hadamard product. Then, summing these scores over proposals, $\alpha_c = \sum_{r=1}^{R} \boldsymbol{\phi_{\scaleto{P}{3pt}}}$, enables the use of a binary cross-entropy loss as image-level loss function:


\begin{equation}
L_{gc}(\alpha_c, y_c) = - \sum_{c=1}^C [(1-y_c) log(1- \alpha_c) + y_c log(\alpha_c)]
\end{equation}

\noindent where $y_c$ is the label indicating the presence or absence of class $c$ in an image. 

Similar to traditional object detectors, we use strong images to compute bounding box regression and classification via the corresponding fully connected layers. We therefore combine weak and strong supervision by providing direct supervision to proposal-level class prediction $\boldsymbol{\gamma_{\scaleto{C}{3pt}}}$. For regression, each bounding box $b$ is parametrised as a four-tuple $(x, y, h, w)$ that specifies its center coordinate $(x, y)$ and its height and width $(h, w)$. For each proposal classified as foreground in a strong image, this regression branch predicts the offset of these coordinates $t^k = (t_x, t_y, t_h, t_w)$. Hence, for every strong image, the following additional loss is computed on a batch of $M$ proposals:

\begin{equation}
L_p (\gamma, u, t, v) = L_{cls}(\gamma,u) + 1[u \ge 1] L_{reg}(t,v)
\end{equation}
\text{where: } \\
\begin{align}
& L_{cls} = - \frac{1}{M} \sum_{r=1}^{M}\sum_{c=1}^{C+1} u_{cr} \log(\gamma_{cr}), 
& L_{reg}(t,v) = \sum_{i \in (x,y,h,w)} \text{smooth}_{L1} (t_i-v_i)
\end{align}


\noindent Parameters $\gamma$ and $u$ constitute the predicted and target proposal classes respectively, $t$ and $v$ the predicted and target bounding box offsets respectively and $ \text{smooth}_{L1}$ is a smooth $L1$ loss function \cite{girshick2015fast}. 

The loss function of the joint detection module is hence $L_{I_{s}}=L_p{+}L_{gc}$ on strong images, while the loss function on weak images is $L_{I_{w}}=L_{gc}$. Enforcing synergy between the two types of supervision regularises the low-shot task thanks to the statistical information provided by weak images. Moreover, due to the instance-level annotations provided by strong images, this also constrains the MIL task and encoder to learn stronger discriminative features between full and partial-extent object proposals.

\noindent\textbf{Online Bounding Box Augmentation Strategy.}
\label{sec3:online_bb_aug}
Learning to update and improve bounding box spatial regions via low-shot regression is highly challenging. When initial inference and ground-truth box overlap is small, large corrections (spatial offsets) are required. Previous work (BCNet~\cite{pan2019low}) actively elects to exclude such challenging samples, further reducing already highly limited data. We alternatively fully exploit available annotations and push our regression branch output through a second forward pass of our OAM (red arrow in Fig. \ref{fig:architecture_overview}).

More specifically; after the first forward pass, we select the $M$ top scoring proposals, per class, corresponding to image-level ground-truth. $M$ is defined as half the size of the proposal batch used to train the strongly supervised component. This accounts for the presence of irrelevant background proposals and allows us to fix this hyperparameter. 
Once regression branch offsets have been applied, our ROI pooling layer ingests the proposals and yields a new set of bounding box features. Loss functions are evaluated using the updated boxes features and combined with the first pass loss.
The overall loss function of our OAM branch is then: $ L_{1B} = L_{I_{s}}^{I} + L_{I_{w}}^{I} + L_{I_{s}}^{II} + L_{I_{w}}^{II}$, where superscripts $I$ and $II$ indicate the first and second pass, respectively. At every iteration, a batch with the same number of weak and strong images is used.

Motivation for our second pass is two-fold. Firstly augmentation is intrinsically provided as new sets of proposal candidates are generated for regression and classification task training. In contrast pre-computed proposals (predominant in WSOD), that lack additional external augmentation strategies, provide only static input, reducing sample variability during training. Secondly, our regression task is regularised as any weak proposals receiving modifications that hinder correct image-level classification are penalised.

\begin{figure*}[t]
\centering
\includegraphics[width=0.95\linewidth]{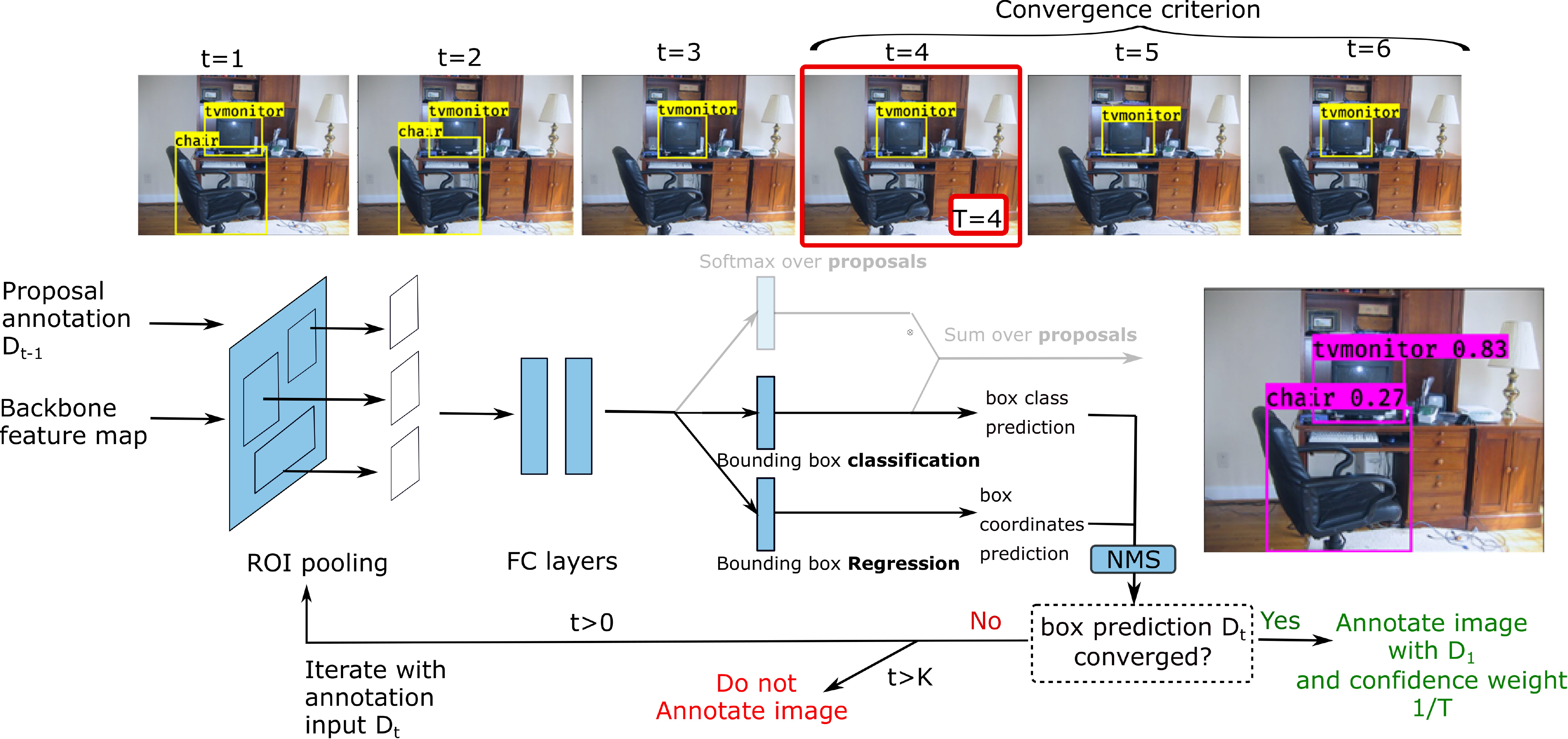}
\caption{Proposed online pseudo-supervision generation strategy. At each iteration, a new set of bounding boxes $D_t$ is computed via classification, regression and NMS of the features from previous set $D_{t-1}$. If bounding box predictions converge at iteration $T<K$, and all proposal classes agree with the image-level label, the weak image is annotated.}
\label{fig:OSG}
\end{figure*}

\noindent\textbf{Online Pseudo-Supervision Generation.}
\label{sec3:online_pseudo_gen}
The key objective of our OAM is to generate reliable annotations on a large set of weakly labelled images in order to guide the training of a fully supervised second branch. As the OAM is trained concurrently with the second branch, it is critical to identify and add only reliable annotations to the pool of training images. Our rationale is that only these images should be used to train the final supervised detection network, while images that the joint detection module struggles to annotate with high confidence should not be used for model training, as they may hurt the training process and deteriorate detector performance.

During early stages of the training process, uncertainty regarding both the class of bounding box proposals and the related regression refinement of box coordinates will be high. As training progresses and model predictive quality improves, confidence, accuracy and stability will increase. This results in an increasingly difficult set of images being accurately annotated. We propose to exploit this behaviour by introducing a supervision generator that is able to reliably identify annotated images, creating a set we refer to as \emph{semi-strong} images $\mathcal{P} \subset \mathcal{W}$, that are used to train the fully supervised branch. Intuitively, $\mathcal P$ will comprise ``easy'' images in early stages of training (\eg single instances, uniform colour backgrounds) and sample diversity will progressively increase as the model becomes more accurate (examples of images annotated by our OAM at different training epochs are reported in Fig. \ref{fig:oa_examples}). 

\begin{figure}[t]
\centering
\includegraphics[width=0.95\linewidth]{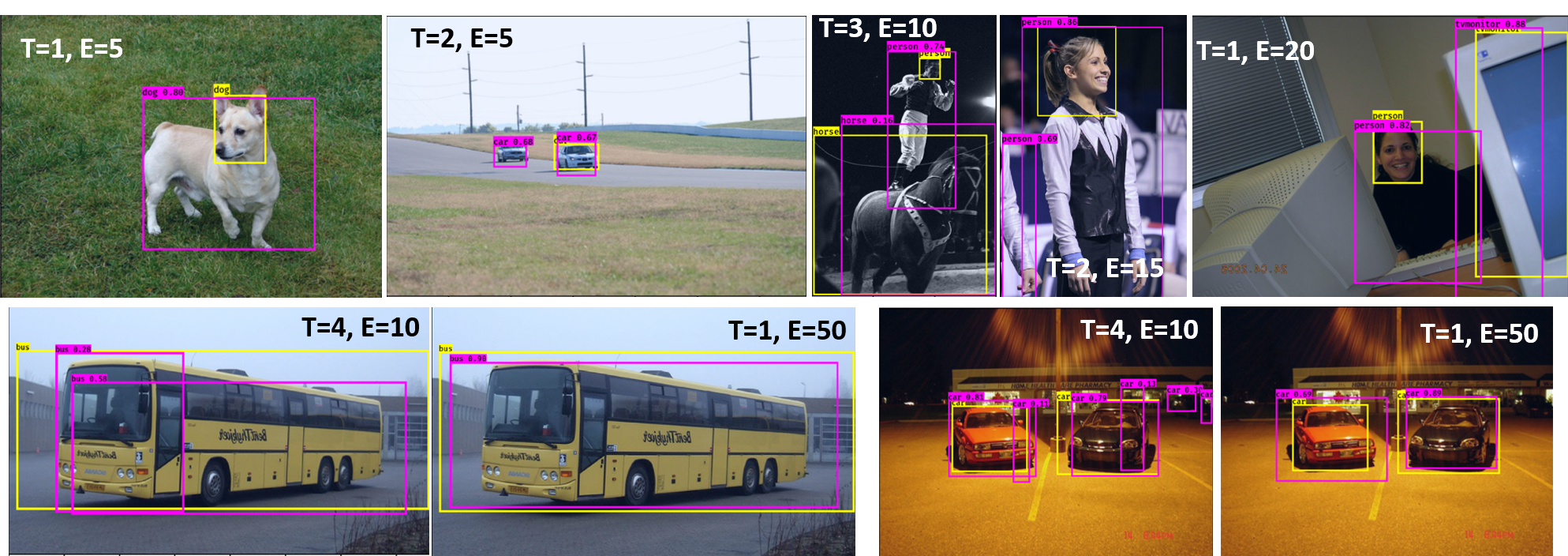}
\caption{Examples of semi-strong images. First row: annotated semi-strong images at epoch $E$, with $T$ iterations required for convergence (see text for details). Second row: examples of semi-strong annotation at pairs of early and late epochs. Magenta color: OAM annotation (class, bounding box score). Yellow: OICR~\cite{tang2017multiple} annotation. The results are obtained from a model trained on PASCAL VOC 2007 with 10 shot strong supervision.}
\label{fig:oa_examples}
\end{figure}

In order to build a set of semi-strong images $\mathcal P$, with bounding boxes and associated annotation confidence scores, we propose the following mechanism. Given a weak image $I$, we obtain a set of $N_1$ bounding boxes $D_1=\{c_r,p_r\}_{r=1}^{N_1}$ after Non-Maximum Supression (NMS) is performed on the output of the joint detection module, where $c_r$ and $p_r$ correspond to the class label and coordinates of box $r$ respectively. 
$D_t=\{c_r,p_r\}_{r=1}^{N_t}$ at every iteration $t>1$, using $D_{t-1}$ as input candidate proposals. More specifically, 
the bounding boxes $D_{t-1}$ obtained at the previous iteration are fed again to the RoI Pooling Layer, providing a new set of image features allowing to compute new proposal coordinates. The process iterates until bounding box prediction stabilises and is stopped when $D_{t}=D_{t-1}$ for three consecutive iterations, i.e. for each bounding box $b_t \in D_t$, there exists a corresponding box $b_{t-1} \in D_{t-1}$ such that $b_t$ and $b_{t-1}$ have intersection-over-union (IoU) $\ge 0.5$ and possess matching class predictions (\ie a standard criterion for characterising object equivalence in detection methods). 
We assign a global confidence weight $1/T$, per image, where $T$ is defined as the first of three iterations in which $D_{t}=D_{t-1}$. Pseudo-code for the OAM algorithm is found in Supplementary Materials A.

The set of proposals $D_1$ obtained at iteration $1$ constitute the final bounding box annotations. Each box is weighted (box level confidence) by its average IoU with the best matching box at all subsequent iterations. Boxes absent at a given iteration (IoU $<0.5$) are, by definition, down weighted due to being assigned an overlap of $0$ at that iteration (Fig.~\ref{fig:OSG} shows an example). Images that do not reach convergence by $K$ iterations, or that fail to find any foreground proposals at iteration $t$, are considered to be annotated with low confidence and are not added to the semi-strong pool. We set the maximum number of updates $K=30$, to prevent large sets of iterations and observe that large $T$ (\eg $T>K$) would only occur during early stage training in practice. Finally, the image is only added to the semi-strong pool 
if the set of obtained annotations contains \emph{all} classes pertaining to the image-level label. 
We highlight that images requiring large iteration count $T$ for convergence are assigned low confidence scores by design and therefore have limited influence on the training procedure of the second branch.  
As weak images get annotated by the proposed OAM during training; the semi-strong set 
expands, while at the same time refining annotations and confidence as the model improves. At a given training step, a weak image that is not successfully annotated, and yet was present in the pool of semi-strong images, will be removed. In this way, the set of semi-strong images 
has the ability to both expand and contract during training.

\subsection{Fully Supervised Branch}
\label{sec3:fully_sup_branch}

Concurrently to OAM training, the obtained strong and semi-strong sets of 
images are used to train a fully supervised second branch, that comprises both bounding box classification and regression modules on the proposal features $\xi_{rf}$ in a similar fashion to Fast(er) R-CNN~\cite{girshick2015fast} style methods. In particular, at every training iteration a batch with the same number of strong and semi-strong images is used. The loss function for this branch is:

\begin{equation}
L_{2B}(p, u, t, v) = L_{cls}(p, u) + L_{reg}(t, v),
\end{equation}

\noindent where $p$ is the ROI class predictions, $t$ is the predicted offset between ROIs and targets, $u$ is the class label and $v$ is the target offset. Only ROIs with foreground labels contribute to the regression loss, $L_{reg}$. The $L_{cls}$ loss constitutes a weighted cross-entropy for each image:

\begin{equation}
L_{cls}(p, u) = - \frac{1}{T} \sum_{i} \omega_i p_i log(u_i)
\end{equation}

\noindent where the proposals in each batch, contributing to the loss, are indexed by $i$, the confidence for GT proposal $u_i$ is denoted $\omega_i$ and the image-level annotation confidence score is denoted $\frac{1}{T}$. Strong images are assigned image and proposal-level weights of $1$. In the early stages of the training process, the semi-strong annotations present some localisation inaccuracies, but are nonetheless highly informative to learn foreground vs background proposals. As training progresses, our OAM improves annotation quality with tighter object coverage and these additional high accuracy annotations will more often contain proposals of exactly full object extent. Such annotations reinforce and strengthen a base signal, provided by strong images alone, towards better bounding-box classification. We also explored utilising semi-strong images to improve bounding-box regression, analogously. In practice, however, this produced slightly worse results. We hypothesise that the 
discrete problem, associated with the bounded classification loss, affords 
more robustness to (early-stage) imperfect semi-strong annotations and therefore compute bounding box regression on only strong images in our final model. To conclude, collecting the introduced components results in the complete loss function for our model: $L_{tot} = L_{1B} + L_{2B}$. At testing, only this fully supervised model is deployed.


\section{Results}
\label{sec4:results}
\subsection{Datasets and Implementation Details}
We evaluate the performance of our proposed method on two common detection benchmarks: the PASCAL VOC 2007~\cite{everingham2015pascal} and the MS-COCO $14$ 
dataset~\cite{lin2014microsoft}, 
referred to as VOC07 and COCO14. 
VOC07 has $5011$ training and $4952$ testing images across $20$ categories. COCO14 has $82k$ training and $5k$ testing images across 80 categories. Following evaluation strategies used in the literature, we evaluate detection accuracy on VOC07 using mean Average Precision (mAP), while we employ the COCO metrics, $AP_{50}$ and $AP_{50:95}$, on the COCO dataset. In the reported experiments, reference to $10\%$ of labelled images dictates that $10\%$ of all images have bounding box annotations while the remaining $90\%$ have image-level labels. This corresponds to $500$ images in VOC07, $8.2k$ images in COCO14. With reference to our ``$N$-shot'' experimental setup, we define each class to have access to $N$ images possessing bounding box annotations. All the experiments on VOC07 use the same data splits provided by BCNet~\cite{pan2019low}, experiments on COCO14 use random selection.

\begin{table*}[t]
\centering
\begin{center}
\scalebox{0.6}{%
\begin{tabular}{ccccccccccccccccccccccc}
\toprule
method & backbone & aero & bike & bird & boat & bottle & bus  & car  & cat  & chair & cow  & table & dog  & horse & moto & person & plant & sheep & sofa & train & tv   & mAP(\%) \\
\toprule
10\% images  &      &      &      &      &      &      &      &      &      &             &      &      &      &      &      &      &      &      &      &      &  \\
\toprule
Fast R-CNN & VGG        &  47.9 & 62.9 & 45.5 & 34.2 & 23.0 & 54.6 & 70.8 & 65.5 & 27.2 & 61.1 & 39.8 & 60.6 & 70.0 & 63.3 & 64.2 & 14.7 & 52.9 & 43.0 & 55.7 & 49.5 & 50.3    \\
BAOD & VGG         & 51.6 & 50.7 & 52.6 & 41.7 & 36.0   & 52.9 & 63.7 & 69.7 & 34.4  & 65.4 & 22.1  & 66.1 & 63.9  & 53.5 & 59.8   & 24.5  & 60.2  & 43.3 & 59.7  & 46.0 & 50.9    \\
BCNet &  VGG       & 64.7 & \textbf{73.1} & 55.2 & 37.0 & 39.1   & \textbf{73.3} & 74.0 & 75.4 & \textbf{35.9}  & 69.8 & 56.3  & \textbf{74.7} & 77.6  & 71.6 & 66.9   & 25.4  & 61.0  & \textbf{61.4} & \textbf{73.8}  & \textbf{69.3} & 61.8    \\
Ours & VGG         & \textbf{65.6} & \textbf{73.1} & \textbf{59.0} & \textbf{49.4} & \textbf{42.5}   & 72.5 & \textbf{78.3} & \textbf{76.4} & 35.4  &\textbf{72.3} & \textbf{57.6}  & 73.6 & \textbf{80.0}  & \textbf{72.5} & \textbf{71.1}   & \textbf{28.3}  & \textbf{64.6}  & 55.3 & 71.4  & 66.2 & \textbf{63.3}    \\
\arrayrulecolor{gray}\midrule\arrayrulecolor{black}
EHSOD &ResNet         & 60.6 & 65.2 & 55.0 & 35.4 & 32.8   & 66.1 & 71.3 & 75.3 & 38.4  & 54.1 & 26.5  & 71.7 & 65.0  & 67.8 & 63.0   & 27.7  & 52.6  & 48.6 & 70.9  & 57.3 & 55.3    \\
BCNet &  ResNet    & \textbf{68.3} & 72.0 & 61.2 & 48.1 & 40.8   & 73.3 & 73.4 & 77.8 & 37.0  & 69.7 & 58.3  & 78.2 & 80.0  & 67.5 & 70.5   & \textbf{27.4}  & 62.9  & \textbf{63.6} & \textbf{73.4}  & 63.6 & 63.4    \\
Ours & ResNet      & 62.3 & \textbf{73.2} & \textbf{61.8} & \textbf{56.2} & \textbf{44.3}   & \textbf{75.4} & \textbf{76.7} & \textbf{80.5} & \textbf{39.5}  & \textbf{73.7} & \textbf{61.7}  & \textbf{78.8} & \textbf{82.8}  & \textbf{71.5} & \textbf{74.3}   & 27.0  & \textbf{67.4}  & 62.7 & 71.2  & \textbf{64.4} & \textbf{65.3}    \\
\toprule
10 shots             &      &      &      &      &        &      &      &      &       &      &       &      &       &      &        &       &       &      &       &      &         \\
\toprule
BCNet & VGG          & 59.7 & 69.1 & 44.6 & 29.4 & 40.1   & 69.2 & 73.2 & \textbf{72.9} & 32.9  & 58.1 & 53.3  & 66.7 & 71.3  & 66.0 & 61.7   & \textbf{24.6}  & 53.0  & \textbf{62.0} & \textbf{67.2}  & \textbf{67.4} & 57.1    \\
Ours & VGG           & \textbf{60.2} & \textbf{71.6} & \textbf{51.5} & \textbf{45.6} & \textbf{43.5}   & \textbf{71.1} & \textbf{75.8} & 72.2 & \textbf{33.8}  & \textbf{62.9} & \textbf{54.0}  & \textbf{70.0} & \textbf{72.9}  & \textbf{67.5} & \textbf{67.4}   & 23.6  & \textbf{61.5}  & 59.1 & 63.6  & 66.7 & \textbf{59.7}    \\
\arrayrulecolor{gray}\midrule\arrayrulecolor{black}
BCNet & ResNet       & \textbf{63.4} & 69.4 & 54.7 & 39.5 & 35.9   & 70.6 & 71.8 & 71.8 & 33.5  & 64.6 & 50.0  & 65.3 & 72.7  & 62.5 & 61.6   & \textbf{29.2}  & 54.5  & \textbf{63.3} & 66.7  & \textbf{69.4} & 58.5    \\
Ours & ResNet        & 61.7 & \textbf{72.3} & \textbf{56.5} & \textbf{52.0} & \textbf{37.2}   & \textbf{71.3} & \textbf{74.6} & \textbf{77.8} & \textbf{36.0}  & \textbf{67.1} & \textbf{58.3}  & \textbf{78.1} & \textbf{77.6}  & \textbf{68.0} & \textbf{71.8}   & 25.5  & \textbf{63.6}  & 62.4 & \textbf{72.7}  & 61.2 & \textbf{62.3}    \\
\toprule
\end{tabular}
}
\end{center}
\caption{Detailed detection performance (\%) on VOC07 dataset. In all the setting, the same BCNet data splits were employed \cite{pan2019low}.}
\label{tab:pascall}
\end{table*}



We employ popular network backbones VGG16 and ResNet101 in our experiments to retain consistency with recent approaches. We combine our OAM with Fast R-CNN \cite{girshick2015fast} (using Edge Boxes~\cite{zitnick2014edge}) and Faster R-CNN using a trainable RPN \cite{ren2015faster}. Optimisation of all models is performed using SGD with weight decay $0.0001$ and momentum $0.9$. For experiments concerning the VOC07 dataset, models are trained for $60$ epochs. The initial learning rate is $0.001$ (first $40$ epochs) and reduced to $0.0001$ for the final $20$ epochs. Analogously for MS COCO experiments; models are trained for 12 epochs, with learning rate $0.001$ in the first 9 epochs and then reduced to $0.0001$ for the final 3 epochs. Remaining model hyper-parameters follow the values reported in~\cite{pan2019low}. For data augmentation, we apply the same augmentation strategy as BCNet~\cite{pan2019low} for fair comparison, \ie we bilinearly resize images to induce a minimum side length $\in \{400,600,750\}$ and, for fully supervised training, uniformly crop image regions with a fixed $600{\times}600$ window. 
All experiments are implemented in PyTorch using a single GeForce GTX 1080 GPU.

\subsection{Comparisons with State-of-the-art}
\textbf{Baselines:}  We evaluate our model with respect to two SOTA WSOD methods, PCL \cite{tang2018pcl} and WSOD$^2$ \cite{zeng2019wsod2}, that were evaluated on both VOC07 and COCO14. We further compare to three MSOD approaches: the two level approach of BCNet~\cite{pan2019low}, end-to-end methods BAOD~\cite{pardo2019baod} and EHSOD~\cite{fang2020low}. To the best of our knowledge, these are the only three methods adopting mixed supervision. All three methods were evaluated on VOC07. 
Results for BCNet, the best performing baseline on VOC07, were not available for the COCO dataset. The approach requires training two models (OICR and BCNet) with two separate sets of parameters that need to be adapted to the new dataset, making it highly challenging and time consuming to provide a fair comparison, hence we were not able to provide it. Similarly, EHSOD was evaluated only on the COCO 2017 database with a much larger set of annotated training images (approx. $12k$), making results not directly comparable to our experiments and different from the low-shot setting studied in this work. Finally, we compare our results with respect to Fast R-CNN and Faster R-CNN trained with full supervision (our upper bounds) and low-shot supervision (\ie $10\%$ and $10$-shot training data), using the same augmentation strategy as all previous models.  

\begin{table*}[t]
\centering
\begin{center}
\scalebox{0.65}{%
\begin{tabular}{|c|c|c|c|c|c|}
\toprule
\hline
Method type & Method               & \multicolumn{2}{c|}{ 10-shots/WSOD} & \multicolumn{2}{c|}{ 10\% images}                   \\
& & \thead{AP (\%) \\  person class} & mAP (\%)& \thead{AP (\%) \\  person class} & mAP (\%) \\
\hline
fully supervised & Fast R-CNN  & 58.0 & 42.1 & 64.2 & 50.3 \\
\rowcolor{black!15} fully supervised & Faster R-CNN  & 54.3 & 37.7 & 55.7 & 46.7 \\
\arrayrulecolor{gray}\hline\arrayrulecolor{black}
WSOD & PCL              & 17.8 & 43.5 & - & - \\
WSOD & PCL + Fast R-CNN & 15.8 & 44.2 & - & - \\
WSOD & WSOD$^2$         & 21.9 & 53.6 & - & - \\
\arrayrulecolor{gray}
\hline
\arrayrulecolor{black}
MSOD & BAOD & - & - & 59.8 & 50.9 \\
\rowcolor{black!15}
MSOD & EHSOD (ResNet + FPN)          & -             & -            & 63.0          & 55.3         \\
MSOD & BCNet                         & 61.7          & 57.1         & 66.9          & 61.8         \\
MSOD & Ours                          & \textbf{67.4} &\textbf{59.7} & \textbf{71.1} & \textbf{63.3}\\
\rowcolor{black!15}MSOD & Ours + RPN & 64.3          & 54.6         & 68.9          & 60.5         \\
\arrayrulecolor{gray}
\hline
\arrayrulecolor{black}
fully supervised & Fast-RCNN 100 \% images (Ours upper bound)         & \multicolumn{4}{c|}{76.8 (person), 71.6} \\
\rowcolor{black!15}
fully supervised & Faster-RCNN 100 \% images (Ours + RPN upper bound) & \multicolumn{4}{c|}{75.6 (person), 67.0} \\
\hline
\bottomrule
\end{tabular}
}
\end{center}
\caption{Comparison to SOTA on VOC07 dataset. A VGG backbone is used unless specified. Gray rows correspond to methods learning an RPN (vs methods using precomputed proposals).}
\label{tab:pasc}
\end{table*}

\noindent \textbf{PASCAL VOC 2007:} 
We report detailed per-class results, compared to competing MSOD approaches in Tab.~\ref{tab:pascall} using 10\% annotated training images, and 10 shots. We consistently outperform all competing methods in terms of mAP, with an improvement of up to $4\%$ with respect to BCNet in the 10 shot scenario (ResNet), and $10\%$ with respect to EHSOD in the 10\% images scenario. We further highlight that BCNet constitutes a two-level WSOD dependent method. The influence of the chosen WSOD component is clearly visible; object classes where their method excels, and surpasses our per-class performance, are the same classes for which their adopted WSOD component (OICR) provides best initial bounding box estimations~\cite{tang2017multiple}. In Tab. \ref{tab:pasc}, we provide more comparisons in the 10 shots and 10\% images scenarios using precomputed proposals (white rows) and an RPN~\cite{ren2015faster} (grey rows). We highlight that we use an off-the-shelf RPN without parameter optimisation, and expect performance to be worse, and not directly comparable to strategies relying on pre-computed proposals. We further compare with top performing WSOD methods and Fast(er)-RCNN approaches and highlight our performance on the ``person'' class, often reported as one of the most challenging classes for WSOD methods due to the large intra-class variability in terms of appearance~\cite{pan2019low,zeng2019wsod2}. We significantly outperform all SOTA methods, and substantially improve with respect to WSOD methods, in particular for the person class, with only minor additional labelling cost. Comparing to Fast(er)-RCNN methods, we highlight that our OAM improves upon models trained on 10\% data and 10 shots by a large margin ($13\%$ and $17\%$ respectively), reaching performance close to the fully supervised upper bound.

\begin{table*}[t]
\centering
\begin{center}
\scalebox{0.65}{%
\begin{tabular}{|c|c|c|c|c|}
\toprule
\hline
Method type & Method                   & AP@.50 & AP@{[}.50,.95{]}  \\ 
\hline
fully supervised & Fast R-CNN  - 10 shots                        &   22.1 &  10.0          \\
\rowcolor{black!15} fully supervised & Faster R-CNN  - 10 shots  &   16.1 &  6.7           \\
\arrayrulecolor{black!25}\hline\arrayrulecolor{black}
WSOD & PCL                       & 19.4   & 8.5            \\
WSOD & PCL+ Fast R-CNN           & 19.6   & 9.2            \\
WSOD & WSOD\textasciicircum{}2   & 22.7   & 10.8           \\
\arrayrulecolor{black!25}
\hline
\arrayrulecolor{black}
                     MSOD & Ours - 10 shots       &  \textbf{31.2}  &  \textbf{14.9}  \\ 
\rowcolor{black!15}  MSOD & Ours + RPN - 10 shots &  \textbf{24.9}  &  \textbf{10.2}  \\ 
\arrayrulecolor{black!25}
\hline
\arrayrulecolor{black}fully supervised & Fast R-CNN  - 100\% data  & 49.9  & 29.0 \\ 
\rowcolor{black!15} fully supervised & Faster R-CNN  - 100\% data  & 42.1  & 20.5 \\ 
\hline
\bottomrule
\end{tabular}
}
\end{center}
\caption{Comparison with the SOTA on MS-COCO14 with 10-shot training examples (VGG backbone).
Gray rows correspond to methods learning an RPN (vs methods using precomputed proposals).}
\label{tab:COCO}
\end{table*}



\noindent \textbf{MS-COCO14:}
We provide further comparison to additional benchmark datasets in order to highlight model generalisability. We note that contemporary WSOD methods mainly focus on detection datasets of modest size such as VOC07. COCO14 is significantly larger, and constitutes a more challenging dataset due to both the increased size and variability expressed in image content. 
Tab.~\ref{tab:COCO} reports comparisons between our method (precomputed and RPN proposals) and WSOD approaches PCL and WSOD$^2$ on COCO14 using 10 shots labelled images. 
As we compare solely to WSOD methods, we limit our experiments to the 10 shots setting, as 10\% annotated examples provide a very significant advantage compared to WSOD methods. 
We additionally provide comparison to Fast(er) R-CNN methods trained on 10 shots as well as their fully supervised equivalent on 100\% images. We highlight that our method maintains robust performance and significantly outperforms the WSOD methods and 10 shots Fast(er)-RCNN models ($9\%$). This provides evidence in support of our claim that the strategy of providing mixed supervision significantly improves generalisation ability in settings that entail more difficult tasks with higher variability.

\begin{table*}[t]
\centering
\begin{center}
\scalebox{0.62}{%
\begin{tabular}{ccc|cc|cccccccccccccccccccccc}
\toprule
\hline
\multicolumn{2}{c}{10 shots}  & \multicolumn{2}{c}{ } & \multicolumn{20}{c}{AP (\%) }\\
\hline
SE &  BBA & OAM & 1B & 2B & aero & bike & bird & boat & bottle & bus  & car  & cat  & chair & cow  & table & dog  & horse & moto & person & plant & sheep & sofa & train & tv   & mAP(\%)  \\
\hline
\rowcolor{black!15} 
& \checkmark & & \checkmark&       &  42.0 & 57.1 & 40.2 & 34.2 & 30.3 & 62.6 & 69.0 & 62.5 & 23.2 & 63.8 & 33.0 & 58.5 & 72.2 & 63.3 & 62.9 & 20.8 & 54.9 & 44.2 & 54.3 & 55.2 & 50.2  \\
&  & &  & \checkmark                       &    30.9 & 53.2 & 35.8 & 27.8 & 19.9 & 51.6 & 65.8 & 54.7 & 19.3 & 48.3 & 27.8 & 46.3 & 57.7 & 54.3 & 58.0 & 14.9 & 49.1 & 37.5 & 43.8 & 44.7 & 42.1   \\
\rowcolor{black!15}
\checkmark & \checkmark & & \checkmark &      &    44.3 & 60.2 & 40.4 & 37.8 & 28.1 & 67.0 & 72.8 & 64.1 & 24.2 & 64.6 & 40.9 & 60.5 & 70.5 & 61.6 & 63.5 & 16.1 & 55.0 & 46.2 & 57.5 & 58.0 & 51.7     \\
\checkmark  & \checkmark & &      & \checkmark  &          47.3 & 62.1 & 42.4 & 35.2 & 28.2 & 67.0 & 72.8 & 65.1 & 21.7 & 65.3 & 43.4 & 61.4 & 70.6 & 63.5 & 63.0 & 16.5 & 57.6 & 45.8 & 58.7 & 54.7 & 52.1     \\
\rowcolor{black!15}
\checkmark  &  & \checkmark &  \checkmark &               &       50.3 & 67.3 & 49.8 & 44.1 & 35.9 & 64.3 & 72.7 & 70.3 & 32.6 & 57.7 & 44.5 & 66.3 & 65.6 & 68.3 & 62.8 & 25.2 & 60.0 & 48.8 & 62.6 & 64.5 & 55.7 \\
\checkmark  &  & \checkmark&  &  \checkmark    &            61.4 & 71.0 & 48.5 & 42.9 & 37.8 & 69.8 & 75.6 & 72.8 & 34.0 & 63.2 & 47.6 & 71.9 & 71.1 & 71.1 & 64.6 & 25.7 & 63.4 & 55.6 & 61.9 & 65.8 & 58.8 \\
\rowcolor{black!15}
\checkmark  & \checkmark & \checkmark & \checkmark  &      & 57.9 & 71.4 & 48.2 & 42.7 & 38.0 & 71.4 & 75.5 & 75.5 & 34.0 & 67.1 & 54.0 & 71.4 & 74.3 & 69.4 & 65.7 & 23.7 & 61.6 & 56.1 & 61.0 & 65.0 & 59.2 \\
\checkmark  & \checkmark & \checkmark &   & \checkmark     & 60.2 & 71.6 & 51.5 & 45.6 & 43.5 & 71.1 & 75.8 & 72.2 & 33.8 & 62.9 & 54.0 & 70.0 & 72.9 & 67.5 & 67.4 & 23.6 & 61.5 & 59.1 & 63.6 & 66.7 & 59.7 \\
\hline
\bottomrule
\end{tabular}
}
\end{center}
\caption{Ablative analysis of our method on VOC07 for the 10 shot scenario. SE: shared encoder, OAM: second branch training also on OAM generated semi-strong images, BBA: bounding box augmentation strategy. 1B: first branch output, 2B: second branch output.}
\label{tab:ablation}
\end{table*}


  


\subsection{Ablation Studies}
\label{sec:as}
We conduct experiments to understand the different contributions and assignment of credit for our OAM components using the VOC07 dataset and a VGG backbone. Tab.~\ref{tab:ablation} shows ablative results for the 10 shots scenario while additional results for the 10\% images scenario are reported in supplementary materials. Studied components are: \emph{SE}: shared encoder (\ie no SE entails independent branch training); \emph{OAM:} fully supervised branch is also trained on semi-strong images generated by the OAM; \emph{BBA:} online bounding box augmentation strategy. For each configuration, we report mAP with respect to the output of the OAM (first branch; 1B) as well as the output of the fully supervised branch (second branch; 2B). We experimentally verify the importance of each component; performance consistently improves as new components are integrated. We note that the shared encoder strongly improves the fully supervised branch, while the OAM, and communication between branches, affords mutual branch improvement. Both performance gains can be attributed to the more discriminative full \vs partial object proposal features learned by the shared encoder.

\section{Conclusion}
\label{sec5:conclusion}
We have introduced a novel online annotation module (OAM), trained using mixed supervision, that learns to generate annotations on the fly and thus affords concurrent training for fully supervised object detection. The OAM can be combined with any two-stage object detector and provides an intrinsic curriculum to improve the training procedure. Extensive experiments on two popular benchmarks show SOTA performance in the mixed supervision scenario, and significant improvement of two-stage detection methods in low-shot settings. Moreover, our method has the potential to increase performance on rare, long tail classes that typically only possess a handful of annotated examples.

\newpage

\begin{subappendices}
\renewcommand{\thesection}{\Alph{section}}%

\chapter*{Appendices}
Here we provide additional material to supplement our work. 

In Appendix \ref{sec:oamp}, we report a pseudocode description of the proposed online pseudo-supervision generation algorithm in the Online Annotation Module (OAM).

In Sec.\ $4.3$ of the main paper, we presented an ablation study to confirm the influence of each component of our method. This was carried out using the $10$ shot scenario, with the PASCAL VOC 07 dataset (VOC07) and a VGG16 backbone. Here in Appendix \ref{sec:ablation_10perc} we present an extended analysis using, alternatively, $10\%$ of VOC07 training images for strong supervision. Additionally, to further explore method sensitivity, Appendix \ref{sec:anno_sens} investigates variance caused by the selection process of the fully annotated image set; we report a five-fold experiment, under the $10$ shot scenario again employing VOC07 with a VGG16 backbone.

The EHSOD paper~\cite{fang2020low} reports detection results for the MS-COCO 17 (COCO17) dataset corresponding to the $10\%$ training data scenario. In that setting, ${\sim}12000$ fully annotated images are available to the model, which strays from the \emph{low-shot} scenario studied in our work. Nonetheless, for completeness, we report comparison between our method (considering both pre-computed and RPN proposal instances) and EHSOD~\cite{fang2020low}, and provide additional qualifying discussion in Appendix \ref{sec:coco17}.

In Appendix \ref{sec:additional_voc07} we report additional detailed per-class detection results for both $20\%$ and $20$ shot annotation scenarios on VOC07, with comparisons to alternative Mixed Supervision Object Detection (MSOD) approaches. Per-class detection results aim to further reader understanding and offer deeper insight into competing methods' performance and individual per-class traits.

In Appendix \ref{sec:add_vis_results} additional visual results are provided; Appendix \ref{sec:add_vis_results:1} and Appendix \ref{sec:add_vis_results:2} show (1) examples of images annotated by our OAM and (2) test time detection performance (for VOC07, COCO14) respectively. Finally, in Appendix \ref{sec:add_vis_results:3}, we highlight some common failure cases of our method.

\section{Online Pseudo-Supervision Generation algorithm}
\label{sec:oamp}

\begin{algorithm}[H]
	
	\caption{Online Pseudo-Supervision Generation algorithm}
	\label{alg1}
	\begin{algorithmic}[1]
	
	    \State{{\bf Input}:
	        Initial set of N detections $D_0=\{c_r,p_r\}_{r=1}^{N}$, 
	        stopping criterion $K$, 
	        image feature vector $f(\bf{x})$, 
	        OAM layers parameters $\theta$. 
	    }
	    \State{{\bf Output}:
	        M output detections  $D_1=\{c_r,p_r,w_r\}_{r=1}^{M}$ with confidence weights $w_r$,
	        number of iterations required for convergence $T$.
	    }
	    \State{\textit{Initialise variables:} $D \gets D_0$, $counter \gets 0$}
	    
	    \State{\bf For} t = 1 {\bf to} K {\bf}:
	    \State \indent $\{  \boldsymbol{\xi_{r}} \}_{r=1}^{N} \gets$ RoIPooling($D,f(\bf{x})$) 
	    \State \indent $D_t$ = $forward_{\theta}(\{  \boldsymbol{\xi_{r}} \}_{r=1}^{N})$
	    
	    \State \indent {\bf if } $D_t$ is empty : \Comment{\normalfont No detections}
	    \State \indent \indent \bf{break}\normalfont
	    \State \indent \bf{if } $D_t == D$  : \Comment{\normalfont $\forall$  $b_t \in D_t,$   $\exists$  $b \in D $ \normalfont where $IOU(b_t,b)\leq 0.5$ and $class(b_t)=class(b)$. }
	    \State \indent \indent  $counter ++$
	    \State \indent \indent \bf{ if } $counter == 3$ :\normalfont
	    \State \indent \indent \indent $T \gets t + 1  - counter$ \Comment{\normalfont First of three iterations where $D_t == D$}
	    \State \indent \indent \indent $w_r \gets $ {\normalfont averageOverlap}$(\{D_t\}_{t=1}^{T})$
	    \State \indent \indent \indent {\bf break}
	    \State \indent \bf{else:} \normalfont
	    \State \indent \indent $counter \gets 0$
	    \State \indent{$D \gets D_t$}

	\end{algorithmic}
\end{algorithm}

\newpage

\section{Ablation study: 10\% data scenario}
\label{sec:ablation_10perc}
In Tab.~\ref{tab:ablation}, we report ablation study results for the proposed model (VGG16 backbone) where $10\%$ of images from VOC07 provide strong supervision. Results for the analogous $10$ shot scenario were reported in the main paper, Sec.\ $4.3$. Considered ablation components are \emph{SE}: presence of shared encoder (\ie no \emph{SE} entails independent branch training); \emph{OAM:} the fully supervised branch is additionally trained on semi-strong images (generated by the OAM); \emph{BBA:} online bounding box augmentation strategy. For each configuration, we report mAP with respect to the output of the OAM (first branch; \emph{1B}) as well as the output of the fully supervised branch (second branch; \emph{2B}). 

As was also observed for the $10$ shot scenario (reported in Sec. $4.3$ of the main paper), the performance increases as additional components are added, providing further evidence for component validity and contribution. The performance gaps between differing ablations are smaller than our analogous main paper experiment due to the increased strong supervision available in the current case. Congruent with the results reported in Sec. \ref{sec:as}, this ablation highlights that the shared encoder strongly improves the fully supervised branch, while the OAM and communication between branches, afford mutual branch improvement.

\begin{table*}[h!]
\centering
\begin{center}
\scalebox{0.57}{%
\begin{tabular}{ccc|cc|cccccccccccccccccccccc}
\toprule
\hline
\multicolumn{2}{c}{10 \%}  & \multicolumn{2}{c}{ } & \multicolumn{20}{c}{AP (\%) }\\
\hline
SE &  BBA & OAM & 1B & 2B & aero & bike & bird & boat & bottle & bus  & car  & cat  & chair & cow  & table & dog  & horse & moto & person & plant & sheep & sofa & train & tv   & mAP(\%)  \\
\hline
\rowcolor{black!15} 
& \checkmark & & \checkmark&       &  56.7 & 69.9 & 52.5 & 42.7 & 36.7 & 72.9 & 76.4 & 70.6 & 31.8 & 72.6 & 48.2 & 66.9 & 77.7 & 68.9 & 67.1 & 22.9 & 59.9 & 55.5 & 62.8 & 63.2 & 58.8  \\
& & &  & \checkmark                       &    47.9 & 62.9 & 45.5 & 34.2 & 23.& 54.6 & 70.8 & 65.5 & 27.2 & 61.1 & 39.8 & 60.6 & 70.& 63.3 & 64.2 & 14.7 & 52.9 & 43.& 55.7 & 49.5 & 50.3   \\
\rowcolor{black!15}
\checkmark & \checkmark & & \checkmark &      &    57.3 & 67.4 & 51.4 & 42.& 37.2 & 72.2 & 77.2 & 72.5 & 31.7 & 69.5 & 52.8 & 71.1 & 76.5 & 67.8 & 67.4 & 21.8 & 57.7 & 54.6 & 64.5 & 62.3 & 58.7     \\
\checkmark  & \checkmark & &      & \checkmark  &          57.5 & 68.2 & 53.& 41.8 & 37.4 & 70.1 & 77.2 & 73.2 & 33.& 69.3 & 54.8 & 71.8 & 78.4 & 69.& 67.7 & 22.2 & 59.4 & 54.3 & 66.1 & 62.3 & 59.3     \\
\rowcolor{black!15}
\checkmark  &  & \checkmark &  \checkmark &               &       64.3 & 69.7 & 56.1 & 48.3 & 39.8 & 71.4 & 78.1 & 76.5 & 37.8 & 71.1 & 56.4 & 76.5 & 76.5 & 70.9 & 68.4 & 25.7 & 62.1 & 55.7 & 70.2 & 65.8 & 62.1 \\
\checkmark  &  & \checkmark&  &  \checkmark    &            67.1 & 70.3 & 56.2 & 48.4 & 42.1 & 71.7 & 76.9 & 76.7 & 39.2 & 71.5 & 60.1 & 74.1 & 79.6 & 71.3 & 70.9 & 26.3 & 61.6 & 56.4 & 71.1 & 66.1 & 62.9 \\
\rowcolor{black!15}
\checkmark  & \checkmark & \checkmark & \checkmark  &      & 66.4 & 71.8 & 57.3 & 50.3 & 41.5 & 72.6 & 78.5 & 77.3 & 38.4 & 71.6 & 59.8 & 74.3 & 79.4 & 71.5 & 71.4 & 26.1 & 61.8 & 57.6 & 72.3 & 66.5 & 63.3 \\
\checkmark  & \checkmark & \checkmark &   & \checkmark     & 65.6 & 73.1 & 59.& 49.4 & 42.5 & 72.5 & 78.3 & 76.4 & 35.4 & 72.3 & 57.6 & 73.6 & 80.& 72.5 & 71.1 & 28.3 & 64.6 & 55.3 & 71.4 & 66.2 & 63.3 \\
\hline
\bottomrule
\end{tabular}
}
\end{center}
\caption{Ablative analysis of our method using VOC07 in the 10\% scenario. \emph{SE}: shared encoder, \emph{OAM}: second branch trained also using OAM generated semi-strong images, \emph{BBA}: bounding box augmentation strategy. \emph{1B}: first branch output, \emph{2B}: second branch output.}
\label{tab:ablation}
\end{table*}

\newpage

\section{Sensitivity to the selected annotation}
\label{sec:anno_sens}
In order to test the sensitivity of our method, with respect to annotated image-subset selection variance, we perform a five-fold experiment, under the $10$ shot scenario. We test using VOC07 and a standard VGG16 backbone architecture. This scenario represents the setting most susceptible and sensitive to image subset selection as the pool of strong images is the smallest among all considered scenarios (including MS-COCO experiments). It can be observed in~\cref{tab:folds} that image selection variance is small. Varying the selected image subset has only minor effect on final mAP, providing evidence towards the robustness of our proposed approach. This variance intuitively reduces further in cases where the model is trained using a larger number of fully annotated images.

\begin{table*}[h!]
\centering
\begin{center}
\scalebox{0.65}{%
\begin{tabular}{c|cccccccccccccccccccc|c}
\toprule
\hline
SPLIT       & aero  & bike  & bird  & boat  & bottle & bus   & car   & cat   & chair & cow   & table & dog   & horse & moto  & person & plant & sheep & sofa  & train & tv    & mAP($\%$) \\
\hline
1         & 64.1& 73.7& 53.0& 49.2& 46.8 & 73.4& 75.1& 70.5& 33.1& 73.4& 46.9& 75.1& 72.4& 69.8& 63.8 & 31.0& 62.6& 52.2& 69.2& 62.5& 60.9  \\
2  & 60.2& 71.6& 51.5& 45.6& 43.5 & 71.1& 75.8& 72.2& 33.8& 62.9& 54.0& 70.0& 72.9& 67.5& 67.4 & 23.6& 61.5& 59.1& 63.6& 66.7& 59.7  \\
3         & 62.5& 73.9& 60.1& 42.0& 40.0 & 74.1& 74.7& 75.2& 33.7& 74.5& 51.4& 71.4& 79.9& 71.9& 64.6 & 30.3& 63.6& 55.8& 64.8& 66.8& 61.6  \\
4         & 62.2& 75.1& 56.1& 42.7& 38.9 & 73.4& 75.3& 75.0& 32.1& 68.1& 46.3& 69.6& 75.3& 71.1& 62.5 & 26.4& 59.3& 54.3& 69.4& 63.4& 59.8  \\
5        & 64.0& 73.5& 60.1& 50.6& 38.9 & 72.6& 75.6& 70.3& 32.7& 70.1& 55.4& 73.9& 75.1& 70.2& 64.3 & 25.6& 62.6& 49.2& 67.9& 65.3& 60.8   \\
\hline
mean           & 62.6& 73.6 & 56.2 & 46.0 & 41.6  & 72.9 & 75.3 & 72.6 & 33.1 & 69.8 & 50.8 & 72.0 & 75.1 & 70.1& 64.5  & 27.4 & 61.9 & 54.1 & 67.0 & 64.9 & 60.6   \\
std         & 1.6  & 1.3  & 3.9  & 3.8  & 3.4   & 1.1  & 0.4  & 2.4  & 0.7  & 4.6  & 4.1 & 2.4  & 2.3  & 1.7  & 1.8  & 3.1  & 1.6  & 3.7  & 2.6  & 1.9  & 0.8 \\ 
\hline
\bottomrule
\end{tabular}
}
\end{center}
\caption{Five-fold experiment for the 10 shot scenario using VOC07 and a standard VGG16 backbone. Fold mean and standard deviation statistics are reported in the final rows. The second split is the split used  in \cite{pan2019low}, and the split used for all our remaining experiments.}
\label{tab:folds}
\end{table*}

\newpage

\section{MS-COCO 2017 comparisons}
\label{sec:coco17}
The EHSOD~\cite{fang2020low} method reported results using the COCO17 dataset, corresponding to a $10\%$ training data scenario. We thus report here comparison between our method (considering both pre-computed and RPN \cite{ren2015faster} proposal setups) and the EHSOD mixed supervision approach. We also provide additional comparison to both Fast and Faster-RCNN methods, trained using the same $10\%$ of COCO17 images, as well as their fully supervised equivalent; using $100\%$ of the training images. Results are found in Tab.~\ref{tab:COCO}. We note this setting corresponds to approximately ${\sim}12000$ fully annotated images, a much larger set than the ones used in all other experiments.

It can be observed that, in this setting, our model performs on-par with EHSOD when using RPN proposals, while significantly outperforms their approach when pre-computed (Edge Boxes) proposals are employed. Moreover, we observe that our method also performs on-par with the Fast(er)-RCNN baselines in the $10\%$ images scenario. Interestingly we note only a reasonably modest gap between Fast(er)-RCNN performance with regard to the considered $10\%$ and $100\%$ baselines. This suggests that the gap between the $10\%$ and $100\%$ setting can be closed by providing the network with images containing object class appearance outliers or by images containing difficult, crowded scenes. As a consequence, the problem, in this setting, can be considered to have a greater affinity with a fully supervised task than with a low-shot setting. This observation provides some explanation towards why our method provides limited improvement in this setting. Images required to improve the detector performance (high information content) may not be annotated with high confidence and therefore not considered for object detector training. As highlighted in our future work discussion (main paper; Sec.\ 5), we believe active learning strategies may prove fruitful in such cases.

\begin{table*}[h!]
\centering
\begin{center}
\scalebox{0.65}{%
\begin{tabular}{|c|c|c|c|c|c|c|}
\hline
Method type & Method                & AP@.50 & AP@{[}.50,.95{]} \\
\hline
fully supervised & Fast RCNN  - 10\% data     &  53.7     &    31.6           \\
\rowcolor{black!15} fully supervised  & Faster RCNN  - 10\% data   &      46.3  &  25.6      \\
\arrayrulecolor{black!25}\hline\arrayrulecolor{black}
\rowcolor{black!15} MSOD &  EHSOD - 10\% data  &   46.8     &    -              \\
\arrayrulecolor{black!25}\hline\arrayrulecolor{black}
MSOD & Ours - 10\% data       & 54.2     & 31.6             \\ 
\rowcolor{black!15} MSOD & Ours + RPN - 10\% data &     46.0   &     25.4            \\
\arrayrulecolor{black!25}\hline\arrayrulecolor{black}
fully supervised  & Fast RCNN  - 100\% data    &      61.6    &    48.0  \\
\rowcolor{black!15} fully supervised  & Faster RCNN  - 100\% data  &  51.1      & 28.8 \\             
\hline
\end{tabular}
}
\end{center}
\caption{Comparison with state of the art on COCO17. All the models were trained with a ResNet101 backbone \cite{he2016deep}, while EHSOD uses FPN \cite{lin2017feature}. Gray rows correspond to methods learning an RPN \cite{ren2015faster} (\vs methods using precomputed proposals). }
\label{tab:COCO}
\end{table*}

\newpage

\section{Additional PASCAL VOC 07 results}
\label{sec:additional_voc07}
We report here detailed \emph{per-class} detection results and compare competing MSOD approaches using both $16\%$ annotated training images and $20$ shot scenarios. Results are found in Tab.~\ref{tab:pascall}. We consistently outperform all competing methods in terms of mAP, with an improvement of up to $4\%$ with respect to BCNet in the 20 shot scenario (ResNet101 \cite{he2016deep} backbone). We highlight that in the $16\%$ training image scenario, we report both EHSOD and BAOD results, trained using $20\%$ of training images as only these results were available. This highlights the ability of our method to outperform these competing models even in the case where we have access to $200$ fewer training examples.

\begin{table*}[h]
\centering
\begin{center}
\scalebox{0.56}{%
\begin{tabular}{ccccccccccccccccccccccc}
\toprule
method & backbone & aero & bike & bird & boat & bottle & bus  & car  & cat  & chair & cow  & table & dog  & horse & moto & person & plant & sheep & sofa & train & tv & mAP(\%) \\
\toprule
$20$ shot            &               &               &               &               &                 &               &               &               &                &               &               &               &                &               &               &               &               &               &               &               &               \\
\toprule
BCNet & ResNet101         & \textbf{66.5} & 67.6          & 56.7          & 40.5          & 40.4            & 72.8          & 71.3          & 76.6          & \textbf{39.4}  & 65.0          & 54.1          & 71.4          & 72.9           & 66.6          & 66.0          & 26.1          & 59.0          & 65.5          & \textbf{67.7} & \textbf{67.6} & 60.7          \\
Ours  & ResNet101         & 66.2          & \textbf{73.3} & \textbf{57.0} & \textbf{53.2} & \textbf{42.8}   & \textbf{76.0} & \textbf{76.0} & \textbf{79.1} & 38.6           & \textbf{74.6} & \textbf{61.1} & \textbf{79.9} & \textbf{77.4}  & \textbf{70.2} & \textbf{73.1} & \textbf{26.7} & \textbf{64.3} & \textbf{65.7} & 67.6          & 64.5          & \textbf{64.4} \\
\toprule
16\% images           &               &               &               &               &                 &               &               &               &                &               &               &               &                &               &               &               &               &               &               &               &               \\
\toprule
BCNet & VGG16          & 63.7          & \textbf{77.2} & \textbf{62.9} & 48.0          & 39.7            & 73.3          & 76.0          & \textbf{78.0} & 39.4           & 72.9          & 56.1          & 75.4          & 79.9           & 69.5          & 70.2          & 31.0          & 60.6          & 62.2          & \textbf{75.0} & 68.6          & 64.0          \\
Ours & VGG16           & \textbf{66.5} & 76.2          & 59.1          & \textbf{53.0} & \textbf{49.2}   & \textbf{77.1} & \textbf{79.4} & 76.9          & \textbf{41.4}  & \textbf{75.4} & \textbf{63.7} & \textbf{80.2} & \textbf{80.9}  & \textbf{71.6} & \textbf{73.0} & \textbf{35.7} & \textbf{67.5} & \textbf{64.0} & 73.5          & \textbf{68.9} & \textbf{66.7} \\
\arrayrulecolor{gray}\midrule\arrayrulecolor{black}

$\text{BAOD}^\star$  & ResNet101    & 57.0          & 62.2          & 60.0          & 46.6          & 46.7          & 60.0          & 70.8          & 74.4          & 40.5           & 71.9          & 30.2           & 72.7          & 73.8          & 64.7          & 69.8          & \textbf{37.2} & 62.9          & 48.4          & 64.1          & 59.1          & 58.6            \\
BCNet & ResNet101         & \textbf{67.3} & 74.2          & 65.2          & 51.7          & 40.8          & \textbf{74.1} & 72.7          & 77.2          & 39.2           & 70.3          & 59.9           & 77.2          & 78.5          & 69.9          & 68.6          & 30.6          & 60.0          & \textbf{68.2} & \textbf{75.9} & \textbf{66.8} & 64.4            \\
$\text{EHSOD}^\star$  & ResNet101   & 65.5          & 72.3          & \textbf{66.7} & 45.6          & \textbf{50.8} & 72.2          & 77.8          & 82.2          & \textbf{44.3}  & 73.1          & 44.8           & 79.3          & 76.0          & \textbf{73.0} & 73.8          & 35.5          & 63.0          & 62.1          & 74.0          & 65.5          & 64.9            \\
Ours & ResNet101          & 65.8          & \textbf{78.8} & 63.7          & \textbf{55.3} & 49.7          & 73.0          & \textbf{79.6} & \textbf{84.5} & 42.7           & \textbf{75.0} & \textbf{61.6}  & \textbf{84.7} & \textbf{83.3} & 71.8          & \textbf{75.1} & 33.9          & \textbf{64.6} & 64.9          & 73.3          & 66.2          & \textbf{67.4}   \\
\toprule
\end{tabular}
}
\end{center}
\caption{Detailed \emph{per-class} detection performance (\%) on VOC07. For each instance of our model, identical data splits, from the BCNet paper~\cite{pan2019low} were consistently used. Method rows marked ${}^\star$ indicate models trained using $20\%$ of images, due to the availability of comparable results, \cf only $16\%$. }
\label{tab:pascall}
\end{table*}

\newpage

\section{Additional visual results}
\label{sec:add_vis_results}

\subsection{Annotated Semi-Strong Images}
\label{sec:add_vis_results:1}

In~\cref{fig:oa_examples} we provide additional examples of images annotated by our OAM, named semi-strong images, during progressive training epochs $E$. These online annotations are obtained by our model using VOC07 data with $10$ shot strong supervision (other examples of semi-strong images are reported in the main manuscript, Fig. 4). We observe that typically uncomplicated and simple images are labelled with high confidence when training begins (for example at epoch rows $E =\{5,10\}$). During later training stages (here $E>10$), more complex images with increased appearance diversity and also with multiple, overlapping object instances are added to the pool by our OAM.  In general, $T$ ranged from 1-10 (first 5 epochs) to 1-3 (end of training); and the semi-strong set contained approx. 10\% (first epochs) to 45-60\% (end of training) of annotated weak images

Furthermore, we compare the annotations obtained by our method (magenta) with annotations generated by a popular Weakly Supervised Object Detection (WSOD) approach; OICR~\cite{tang2017multiple} (yellow detections). We highlight that, from early epochs, our method is providing better, more reliable annotations that are then employed for concurrent object detector training. Moreover, our annotations cover the full extent of the object of interest. This can be explained due to the high quality information being distilled from the low-shot fully annotated images (strong images), while the WSOD method annotations exhibit the well understood problem of tending to focus on object parts and on (only) the most discriminative object in the image.

\begin{figure}[h]
\centering
\includegraphics[width=1\linewidth]{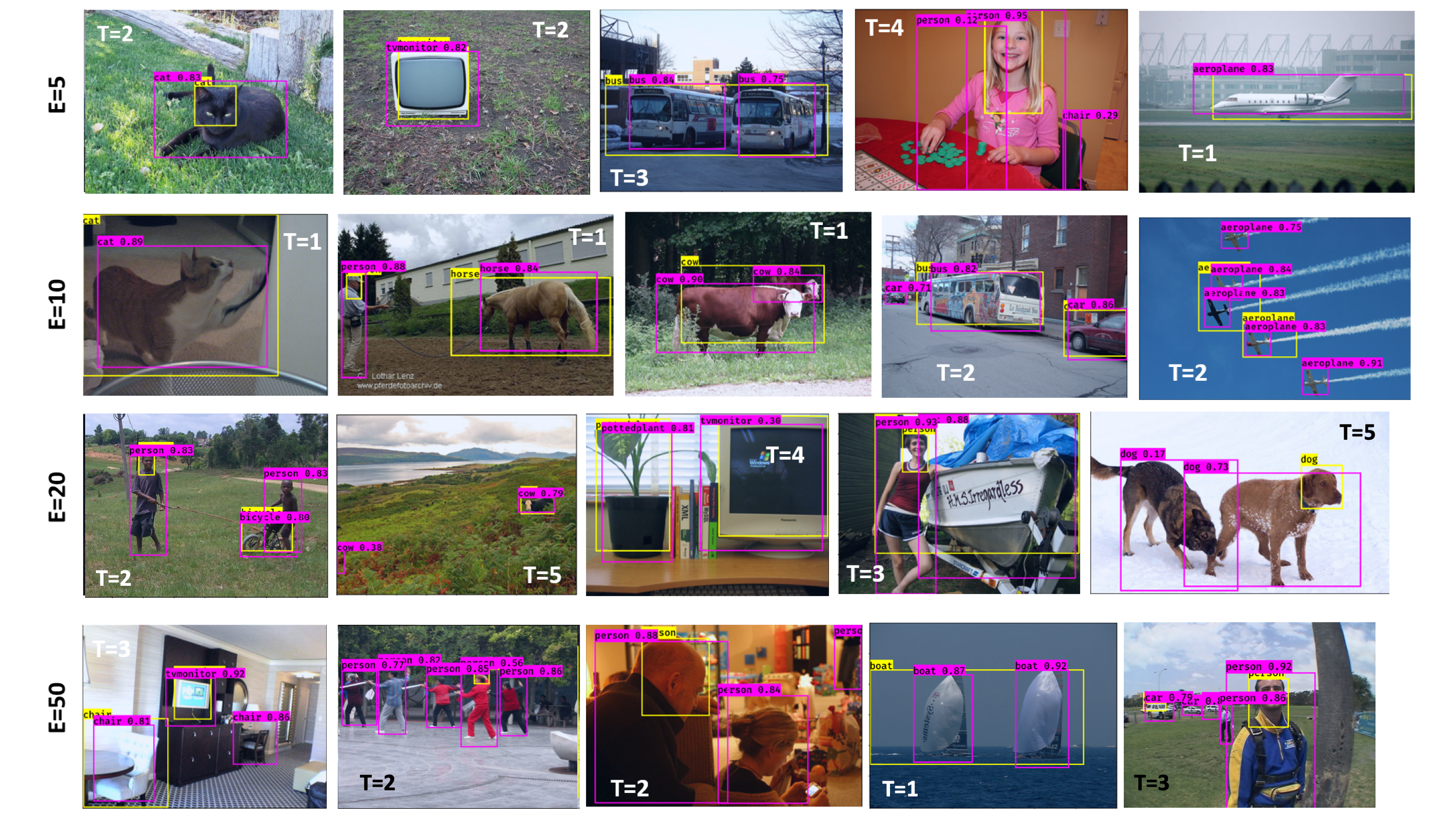}
\caption{\small{Examples of semi-strong images at epoch $E$ with iterations required for OAM convergence $T$ (definition in the main paper, Sec. 3). Magenta: our OAM annotation (class, bounding box score). Yellow: OICR~\cite{tang2017multiple} (WSOD) annotations. Results are obtained using model trained on VOC07 with $10$ shot strong supervision.}}
\label{fig:oa_examples}
\end{figure}

\clearpage
\newpage
\clearpage

\subsection{Examples of Detections}
\label{sec:add_vis_results:2}

\begin{figure}[h!]
\centering
\includegraphics[width=\linewidth]{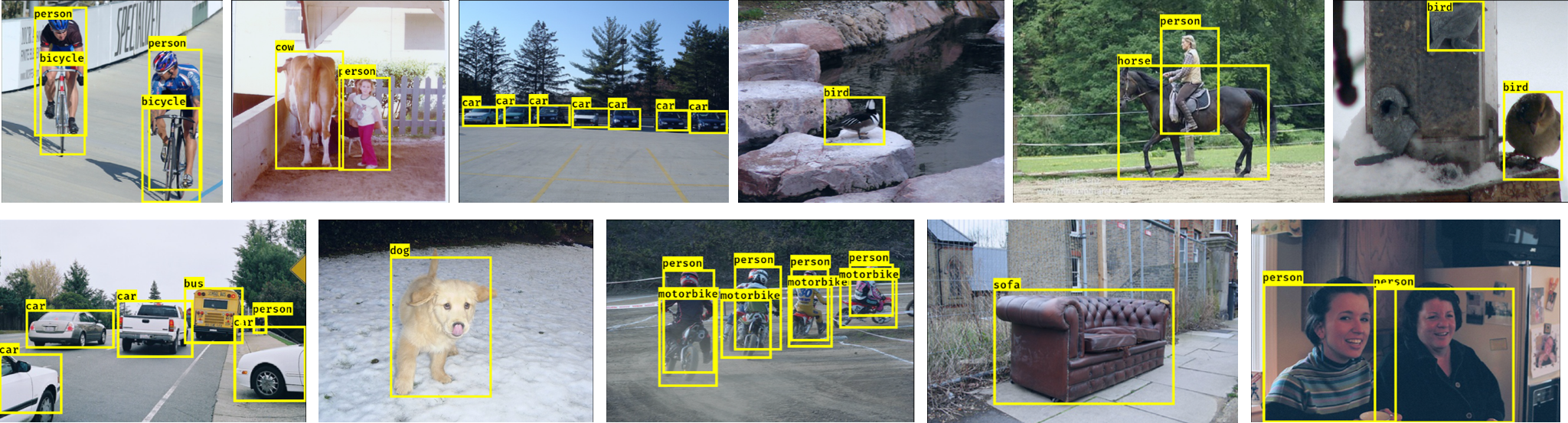}
\caption{Detection results on VOC07 test. Results are obtained from a model trained on VOC07 with $10$ shot strong supervision, VGG16 backbone.}
\label{fig:pascal_res}
\end{figure}

\begin{figure}[h!]
\centering
\includegraphics[width=\linewidth]{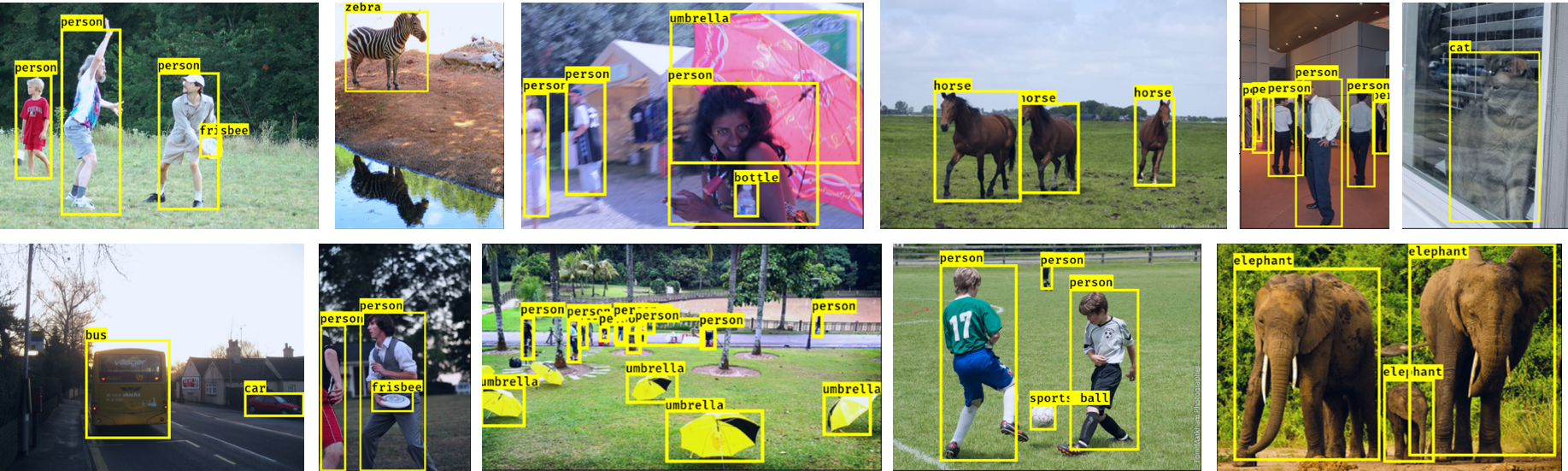}
\caption{Detection results on COCO14 test. Results are obtained from a model trained on COCO14 with $10$ shot strong supervision, VGG16 backbone.}
\label{fig:coco_res}
\end{figure}

Further exemplar test-time detections, obtained by our method with $10$ shot strong supervision, are shown in~\cref{fig:pascal_res} and~\cref{fig:coco_res} for VOC07 and COCO14 test images respectively. Due to the low-shot set of fully annotated images, that are leveraged by our model, we observe that obtained detections cover full object extent, even for classes typically difficult for WSOD (\eg \emph{person}). In comparison with WSOD approaches, our method avoids enclosing only the most discriminative object parts.  Moreover, multiple instances of the same class within a single image can now be captured. This is usually problematic when training a model by relying only on image-level supervision, as in WSOD.

\clearpage
\newpage
\clearpage

\section{Common Modes of Failure}
\label{sec:add_vis_results:3}

We conducted additional investigation to identify instances of detection failures for our model trained with $10$ shot supervision. For both datasets (VOC07, COCO14) considered in our work, the most common mode of failure is represented by multiple detection for an object of interest. Given that the model is only trained with $10$ shot, we partially attribute such failures to the (weakly-learned) bounding box regressor. In corroboration with competing work~\cite{pan2019low,fang2020low} we note bounding box regression is an intrinsically difficult task, especially in cases when limited training data is available or where substantial background pixels need be included to provide an optimal object bounding box, such as for objects with elongated or articulated shapes. As discussed in the main paper (Sec. 5), additional future work may explore strengthening of regression task performance.

\begin{figure}[h]
\centering
\includegraphics[width=0.9\linewidth]{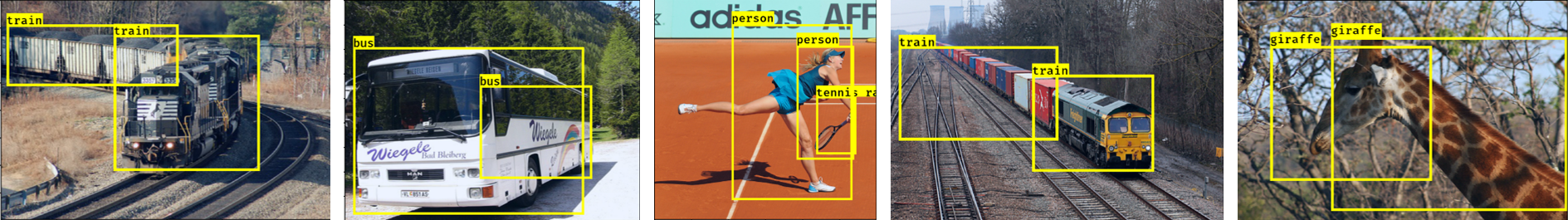}
\caption{Example detection failures obtained from our proposed model. Images are obtained from a model trained on VOC07 (left-most two images) and on COCO14 (right-most three images) with $10$ shot supervision. }
\label{fig:oa_failure_cases}
\end{figure}

\newpage

\end{subappendices}

\bibliographystyle{splncs04}
\bibliography{egbib}
\end{document}